\documentclass[sigconf]{acmart}

\renewcommand\footnotetextcopyrightpermission[1]{} 
\AtBeginDocument{%
  }

\newcommand{\update}[1]{\textcolor{black}{#1}}

\setcopyright{acmlicensed}
\copyrightyear{2018}
\acmYear{2018}
\acmDOI{XXXXXXX.XXXXXXX}
\acmConference[Conference acronym 'XX]{Proceedings of XX}{Date, 2026}{Woodstock, NY}
\acmISBN{978-1-4503-XXXX-X/2018/06}
\usepackage{algorithmic}
\usepackage{algorithm}
\usepackage{graphicx}
\usepackage{caption}
\usepackage{subcaption}
\usepackage{caption}
\usepackage[table]{xcolor}
\usepackage{arydshln}

\usepackage{amsthm}
\usepackage[english]{babel}

\theoremstyle{remark}

\newtheorem{assumption}{Assumption}

\begin{document}

\title{LUMI: Unsupervised Intent Clustering \\with Multiple Pseudo-Labels}


\author{I-Fan Lin}
\affiliation{%
  \institution{Leiden University}
  \city{Leiden}
  \country{The Netherlands}}
\email{i.lin@liacs.leidenuniv.nl}

\author{Faegheh Hasibi}
\affiliation{%
  \institution{Radboud University}
  \city{Nijmegen}
  \country{The Netherlands}}
\email{faegheh.hasibi@ru.nl}

\author{Suzan Verberne}
\affiliation{%
  \institution{Leiden University}
  \city{Leiden}
  \country{The Netherlands}}
\email{s.verberne@liacs.leidenuniv.nl}

\renewcommand{\shortauthors}{Trovato et al.}

\begin{abstract}
In this paper, we propose an intuitive, training-free and label-free method for intent clustering in conversational search.
Current approaches to short text clustering use LLM-generated pseudo-labels to enrich text representations or to identify similar text pairs for pooling. The limitations are: (1) each text is assigned only a single label, and refining representations toward a single label can be unstable; (2) text-level similarity is treated as a binary selection, which fails to account for continuous degrees of similarity. Our method LUMI is designed to amplify similarities between texts by using shared pseudo-labels. 
We first generate pseudo-labels for each text and collect them into a pseudo-label set. Next, we compute the mean of the pseudo-label embeddings and pool it with the text embedding. Finally, we perform text-level pooling: Each text representation is pooled with its similar pairs, where similarity is determined by the degree of shared labels. 
Our evaluation on four benchmark sets shows that our approach achieves competitive results, better than recent state-of-the-art baselines, while avoiding the need to estimate the number of clusters during embedding refinement, as is required by most methods.
Our findings indicate that LUMI can effectively be applied in unsupervised short-text clustering scenarios. Our source code is available at \url{https://anonymous.4open.science/r/pseudo_label-7AE1/README.md} 

\end{abstract}



\received{20 February 2007}
\received[revised]{12 March 2009}
\received[accepted]{5 June 2009}

\maketitle

\section{Introduction}

\begin{figure}[t]
        \centering
        \includegraphics[trim={0 1.2cm 0 0},clip,width=\columnwidth]{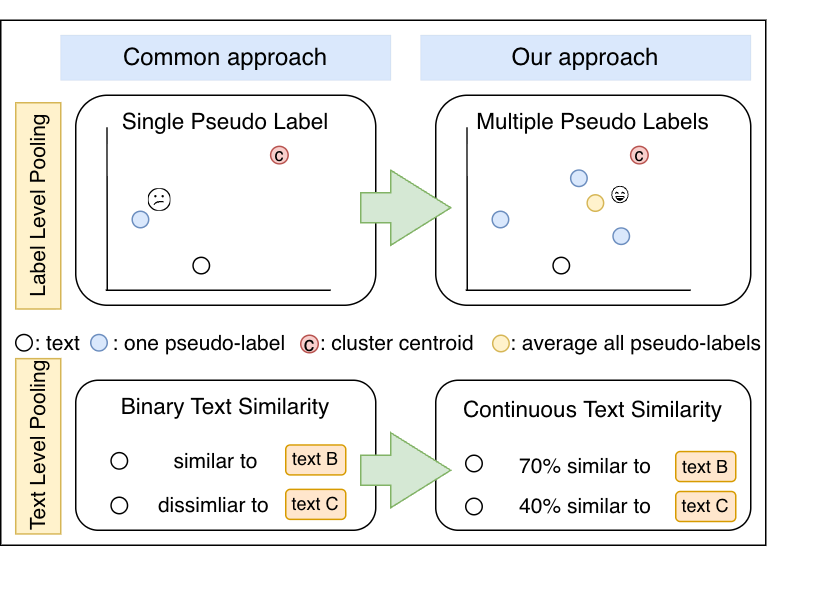}
        \caption{Illustration of our proposed method. Label-level pooling: pooling with multiple pseudo-labels. 
        Text-level pooling: similar pairs are obtained via soft selection based on shared pseudo-labels. 
        }
        \label{fig:idea}
\end{figure}

Intent clustering groups unlabeled user utterances in conversational systems into clusters of similar intents. 
This facilitates query intent identification~\cite{borvcin2024optimizing,alexander2025few,hong2016accurate}, enabling the development of intent-aware information-seeking systems~\cite{koskela2018proactive,qu2019user} 
and customer-facing conversational agents~\cite{liang2024survey}. 

Many studies have addressed this task, mostly aiming at fine-tuning an embedder to learn representations for clustering~\cite{liang2024actively,zou2025glean}. These contrastive learning approaches generally require labeled training data, and their loss functions tend to be complex, making hyperparameter optimization difficult. LLM-guided label-free approaches have been proposed to address these challenges, either guiding representation refinement using LLM-generated pseudo-labels \cite{viswanathan-etal-2024-large,de2023idas}, or by performing similar-text pooling based on LLM judgments \cite{lin2025spill}. The LLM-guided approach performs on par with contrastive learning but with two main limitations: (1) noisy pseudo-labels due to single-shot labeling, and (2) selection restricted to binary decisions, which ignores continuous similarity.

To address these limitations, we propose a method, LUMI, that \textbf{constructs a pseudo-label set to enhance text representations}. Figure \ref{fig:idea} shows the general idea of our approach. Each text is assigned multiple temporary, LLM-generated labels, which serve as cluster-level signals. These pseudo-labels are used in two ways: first, their embeddings are averaged to obtain a stable representation (label-level pooling), and second, they define similarity between texts for further pooling with similar texts (text-level pooling).

Our paper makes three contributions: (1) we theoretically justify why multi-labeling improves clustering, (2) we empirically show that LUMI outperforms state-of-the-art baselines and remains stable across different LLMs and embedders, and (3) our approach enables training-free, and label-free embedding refinement for effective intent clustering, without requiring the number of clusters during refinement.

\section{Related Work}

Several approaches have explored the use of LLMs to improve representations for short text clustering. We categorize them into two types: LLM-as-a-matcher and LLM-as-a-labeler: In the \textbf{LLM-as-a-matcher} setting, 
the LLM is used only to select similar candidates. Zhang et al.~\cite{zhang2023clusterllm} use LLMs to select quality triplets for fine-tuning an embedder. Feng et al.~\cite{feng2024llmedgerefine} use LLMs to re-cluster edgepoints. Lin et al.~\cite{lin2025spill} first encode all texts using an existing embedder and then use an LLM to select similar candidate texts for each instance; the embeddings of these candidates are average-pooled to produce the final representation for clustering. In contrast, the \textbf{LLM-as-a-labeler} approach directly uses LLMs to generate (pseudo-)labels for texts. De Raedt et al.~\cite{de2023idas} first select prototype texts and use an LLM to generate their labels. They then use the LLM again to label/generate the remaining texts with these prototypes. The enriched texts are subsequently used to improve representations. Viswanathan et al.~\cite{viswanathan2024large} use LLMs to generate key phrases for each text, which are then separately encoded into embeddings and concatenated for clustering. Unlike previous work, LUMI is a hybrid method that bridges LLM-as-a-matcher and LLM-as-a-labeler via multiple pseudo-labeling.

\section{Theoretical Foundations}

\update{We provide a theoretical analysis of the label-level pooling here, noting that it constitutes only one component of our overall method.}

\subsection{Problem Formulation}
Following established formulations of emerging intent discovery~\cite{zhang2023clusterllm,viswanathan-etal-2024-large,lin2025spill} we assume
 a collection of text data, denoted as $D = \{x_{i}\}_{i=1}^{N}$, where each data point $x_{i} \in D$ corresponds to a short text, and $N$ is the total number of data points. The task is to partition $D$ into $K$ cluster sets, denoted as $\{S_l\}^K_{l=1}$. Our approach does not rely on a pre-defined $K$ to refine the representation of data points. 
 This is unlike previous studies that involve known $K$ in their method~\cite{de2023idas,zhang2023clusterllm,feng2024llmedgerefine}.

\subsection{Theoretical Justification}%

In this section, we provide a theoretical justification for why label-level pooling improves clustering performance. We adopt the standard clustering assumption that a cluster $S$ is good if its points are tightly concentrated, i.e., the distances between the individual data points and the centroid of the cluster are small.


For an arbitrary cluster, Let $x$ be a text from the cluster $S$ and $\vec{X} = [X_1, \ldots, X_d]$ denote its embedding, where each entry $X_h$ is the $h$-th dimension of the text embedding. Let $\vec{Y}_j = [Y_{j1}, \dots, Y_{jd}]$, for $j = 1, \dots, M$, denote the embeddings of $M$ pseudo-labels of the text, where $Y_{jh}$ is the $h$-th dimension of the $j$-th pseudo-label embedding. Let $c_h$ be the $h$-th dimension of the cluster centroid of the text.

It has been shown empirically that the use of one pseudo-label improves unsupervised clustering ~\cite{de2023idas, viswanathan-etal-2024-large}. We theoretically prove that the average pseudo-label embeddings $\bar{Y} :=\frac{1}{M}\sum_{j=1}^M \vec{Y}_j$ are closer to the cluster centroid than individual pseudo-label embedding $\vec{Y}_j$, which will lead to a better clustering result. Formally, the goal is to ensure the following inequality holds:

\begin{equation}\label{obj_fun}
\sum_{h=1}^{d} E[( \bar{Y}_{h}- c_h)^2] < \sum_{h=1}^{d}E[(Y_{jh} -c_h)^2] 
\end{equation}
where $d$ is the embedding dimension.
%
For notational simplicity, we consider $h$ as a representative dimension and omit the dimension index $h$ from Eq.~\eqref{obj_fun}, rewriting it as:
\begin{equation}\label{obj_fun_one_term}
 E[( \bar{Y} - c)^2] < E[(Y_{j} - c)^2].
\end{equation}
Note that if Eq.~\eqref{obj_fun_one_term} holds, meaning that each dimension of $\bar{Y}$ has a lower mean squared error than the corresponding dimension in $\vec{Y}_j$, then Eq.~\eqref{obj_fun} holds. The inequality \ref{obj_fun_one_term} is equivalent to:
\begin{equation}\label{var_bias}
 \text{Var}(\bar{Y}) + \text{Bias}(\bar{Y}, c)^2< \text{Var}(Y_j) + \text{Bias}(Y_j, c)^2.
\end{equation}

\noindent To prove Eq.~\eqref{var_bias}, we make the following assumptions about all pseudo-labels generated for the text $x$:

\begin{assumption}
    The random variable $Y_j$, representing a dimension of the a pseudo-label embedding for text $x$,  is decomposed as:
\begin{equation} \label{modeling_y}
Y_{j} = c + \alpha U + \epsilon_{j} ,
\end{equation}
where $U := X - c$ is the text-specific deviation, $\alpha \in \mathbb{R}$ controls its contribution, and $\epsilon_j \in \mathbb{R}$ is random noise from the label generation process. This assumption also implies that the LLM-generated pseudo-labels for a randomly selected text $x$ are derived from the same input $x$ (for analysis simplicity).
\end{assumption}

\begin{assumption}
    The i.i.d. variables  $\{\epsilon_{j}\}_{j =1}^{M}$ are independent of $U$.
\end{assumption}

\noindent 
This assumption implies that variables  $\{\epsilon_{j}\}_{j =1}^{M}$, driven from the same distribution, have the same variance $\sigma^2_{\epsilon}$. Under the assumption 1, $\text{Bias}(\bar{Y}, c)^2 = \text{Bias}(Y_j, c)^2$ in Inequality~\eqref{var_bias}. Therefore, it remains only to show $\text{Var}(\bar{Y}) < \text{Var}(Y_j).$ We write the variance of $Y_j$ and $\bar{Y}$ as: 
\begin{align}
    \text{Var}(Y_{j}) &= \alpha^2\sigma^2_{U} + \sigma^2_{\epsilon} \label{single_var}, \\
    \text{Var}(\bar{Y}) &= \alpha^2\sigma^2_{U} +\frac{1}{M} \sigma^2_{\epsilon} \label{muti_var},
\end{align}
where $\sigma^2_{U}$ is variance of $U$. Based on Eqs.~\eqref{single_var} and~\eqref{muti_var}, inequality~\eqref{obj_fun_one_term} and therefore inequality~\eqref{obj_fun} hold. This shows that averaging multiple pseudo-labels reduces noise from the label generation process. Empirically, pseudo-label embeddings also tend to reduce text-specific variability ($\alpha$), which further improves clustering, though this effect is not assumed in the analysis. Note that  in LUMI, pseudo-labels for a given $x$ can be derived from other $x$’s, which often improves label quality.

\begin{algorithm}[t]
\caption{Derivation of Averaged Pseudo-label Embeddings}
\small   
\label{alg:meta_label}
\begin{algorithmic}[1]

\REQUIRE $D $, $D_l$, embedder $f$, candidate size $b$, LLM

\ENSURE Averaged Pseudo-Label Embeddings of the dataset $D$ $\mathbf{Z} \in \mathbb{R}^{N \times d}$

\

\STATE Initialize Averaged Pseudo-Label Embeddings: $\mathbf{Z} \gets \mathbf{0} \in \mathbb{R}^{N \times d}$

\STATE Initialize pooled embeddings: $\mathbf{P} \gets \mathbf{0} \in \mathbb{R}^{N \times d}$  


\FOR{$x_i \in D, l_i \in D_l$} 
    
    \STATE $\mathbf{p}_i \gets 
    \text{Normalize}\!\left( \frac{f(x_i) + f(l_i)}{2} \right)$
    \COMMENT{\# Text-label embedding}
    \STATE $\mathbf{P}[i] \gets 
    \mathbf{p}_i$
\ENDFOR

\FOR{$x_i \in D, l_i \in D_l$}
    
    \STATE $M_i \gets \{(x_{i1}, l_{i1}), \dots, (x_{ib}, l_{ib})\}$ ~~
    \COMMENT{\# Top-$b$ closest texts with a different initial pseudo-label using embeddings $\mathbf{P}$}
    
    \STATE $O_i \gets \text{LLM}(x_i, M_i)$ ~~
    \COMMENT{\# Performing Multi-label classification to get LLM selected pseudo-label set for text $x_i$. $0 \leq |O_i| \leq b$}
    
    \STATE $L_i \gets \{l_i\} \cup O_i$ ~~\COMMENT{\# Final pseudo-label set for $x_i$}

    \STATE $\mathbf{z}_i \gets 
    \text{Normalize}\bigg (\frac{1}{|L_i|}
    \sum_{l \in L_i} f(l)\bigg )$ 
    
    \STATE $\mathbf{Z}[i] \gets 
    \mathbf{z}_i$ ~~\COMMENT{\# text $x_i$ Averaged Pseudo-Label Embedding}
\ENDFOR

\RETURN $\mathbf{Z}$

\end{algorithmic}
\end{algorithm}

\section{Computational Method: LUMI}
Our method has three main steps: (1) generating a single initial pseudo-label for each text, (2) performing multi-label classification to assign multiple pseudo-labels to each text; and (3) using the resulting pseudo-labels for label-level and text-level pooling to obtain a refined embedding for each text.
\subsection{Generation of the pseudo-labels with LLM} For each text $x_i \in D$, we derive a single pseudo-label by prompting an LLM. To ensure labels reflect cluster-level semantics, we provide the LLM with the top $b$ closest texts from the collection based on cosine similarity, where the embeddings are obtained from an existing pre-trained embedder. In our experiments, we set candidate size $b = 10$, since larger $b$ will increase the LLM inference cost. Because the pre-trained embedder is not optimized for our dataset, we apply a simple adaptation: we prompt the LLM to generate labels based only on neighbors that closely match the text. We denote $l_i$ as the initial pseudo-label of $x_i$, and $D_l =[l_1,...,l_N]$ be the ordered list of pseudo-labels corresponding to the dataset $D$.

\subsection{Label-level Pooling} To get pseudo-labels, we perform multi-label classification for each text. For a text $x \in D$, we provide the LLM with the top-$b$ ($b = 10$) nearest neighbors, each associated with a different pseudo-label. The embeddings for selecting these neighbors are computed by pooling the text with its initial pseudo-label, a strategy shown to improve embeddings, even when only a single pseudo-label is used \cite{de2023idas,viswanathan-etal-2024-large}. Once the LLM assigns new pseudo-labels, we encode them separately and pool them together with the initial label to form the averaged pseudo-label embedding, denoted as $\mathbf{z}$. \update{Algorithm~\ref{alg:meta_label} details how $\mathbf{z}$ (uppercase $\mathbf{Z}$ for all texts in $D$) is computed. We then follow prior work \cite{de2023idas} and compute the average of the text embedding $f(x)$ and $\mathbf{z}$, denoted as $\mathbf{z'} = \frac{f(x) + \mathbf{z}}{2}$ (uppercase $\mathbf{Z'}$ for all texts in $D$)}

\subsection{Text-level Pooling}

Once we obtain the embeddings $\mathbf{Z}'$, which incorporate the pseudo-label level information, we use them to perform text-level pooling. Lin et al. \cite{lin2025spill} have conceptually shown that pooling with similar texts shrinks clustering variance and highlight the importance of identifying similar text pairs. However, existing approaches \cite{de2023idas, lin2025spill} treat similarity as binary, without accounting for the varying degrees of similarity; we address this by using a continuous similarity–based pooling. Motivated by the intuition that texts sharing more pseudo-labels should be considered more similar, we propose to use the Jaccard similarity to weight similar texts according to the extent of their shared labels when performing pooling over $\mathbf{Z'}$.

\begin{equation*}
w_{ij} = \frac{|L_i \cap L_j|}{|L_i \cup L_j|},
\end{equation*}
where, $L_i, L_j$ denote the updated sets of pseudo-labels $x_i$ and $x_j$ after multi-label classification. Then, we update:

\begin{equation*}
\tilde{\mathbf{z}}'_i = \frac{\sum_{x_j \in D} w_{ij} \mathbf{z}'_j}{\sum_{x_j \in D} w_{ij}}, 
\end{equation*}

where, $\mathbf{z}'_j$ is the embedding of $x_j$ after averaging the text embedding $\mathbf{x}_j$ with its averaged pseudo-label embedding $\mathbf{z}_j$ and $w_{ii} = 1$. The refined embedding $\tilde{\mathbf{Z}}$ will be used to form clusters.

\section{Experiments and Results}

\subsection{Experimental Setup}

\textbf{Datasets and models} Following most prior work~\cite{zhang2023clusterllm,viswanathan-etal-2024-large,feng2024llmedgerefine,rodriguez2024intentgpt,lin2025spill}, we use Bank77~\cite{Casanueva2020}, CLINC150~\cite{larson2019evaluation}, Mtop~\cite{li2020mtop}, and Massive~\cite{fitzgerald2023massive} for our experiments. We adopt the same example sets: 3,080 examples with 77 intents from Bank77, 4,500 examples with 150 intents from CLINC150, 4,386 examples with 102 intents from Mtop, and 2,974 examples with 59 intents from Massive. Bank77 is single-domain with fine-grained intents, CLINC150 is multi-domain (10 domains), and MTop and MASSIVE are imbalanced datasets. We use all-MiniLM-L6-v2 \citep{reimers-2019-sentence-bert} and Instructor-large \cite{su2023one} as the backbone embedders as they are commonly used in prior work. We use Gemma-2-9b-it~\cite{riviere2024gemma} and Llama3.1-8B-Instruct~\cite{llama3modelcard} as the LLMs.

\textbf{Baseline Methods} KeyphraseCluster \cite{viswanathan2024large} and SPILL \cite{lin2025spill} are directly comparable to LUMI since we do not use the number of clusters $K$ in the representation refinement stage. In contrast, this prior knowledge of the number of clusters is required in ClusterLLM \cite{zhang2023clusterllm} and LLMedgeRefine \cite{feng2024llmedgerefine} to refine the embeddings. For IDAS \cite{de2023idas}, we reproduce its known $K$ and unknown $K$ scenarios. 


\textbf{Evaluation} Following most prior work \cite{de2023idas,zhang2023clusterllm,viswanathan-etal-2024-large,lin2025spill,zou2025glean}, we apply K-means, the most widely used and standard approach for deriving clusters from the constructed embeddings. For evaluation purposes against the labeled test set, the number of clusters is set to the number of ground-truth labels, $K$, as in all prior work, to ensure a fair comparison. After obtaining the clusters, we evaluate them using standard clustering metrics: normalized mutual information (NMI) and clustering accuracy (Acc) \cite{rand1971objective,meilua2007comparing,huang2014deep,gung2023intent}.

\subsection{Main results} 

Table \ref{tab:main_result} shows the comparison between our approach and other SOTA baselines by embedders and LLMs. First, it shows that LUMI is consistently better with both backbone embedders than all other approaches. This implies two points: First, LUMI is more useful and practical than the prior methods IDAS, LLMEdgeRefine, and CLusterLLM, which use the cluster $K$ information in the embedding refinement; omitting the ground-truth $K$ in IDAS leads to a drop in performance. Second, we obtain a stronger result than methods using GPT3.5 (by 10-13 percentage points on ACC) and GPT4 (by 0-3 percentage points on ACC), which means that LUMI reduces the reliance on large-size models. Analyzing results by dataset, our approach outperforms SOTA baselines on three of the four datasets, with the exception of Bank77, where performance fluctuates relative to other SOTA baselines (see Section \ref{banking_error} for discussion).

\begin{table*}[t]
  
  \caption{Results on 4 benchmark sets (averages over 5 runs). \textbf{Bold} = best within each embedder. \underline{Underlined} = best within each LLM. $^*$ = statistically significant vs. strongest baseline (IDAS$\dagger$) on the same dataset within each LLM and embedder. An $\dagger$ indicates that the method incorporates known $K$ in the embedding refinement. IntentGPT, LLMEdgeRefine, KeyphraseClust. and ClusterLLM results cited from prior work.}
  \label{tab:main_result}
  \small  
  \centering
    \begin{tabular}{l|cc|cc|cc|cc|cc}
    \hline
    &\multicolumn{2}{c|}{Bank 77} & \multicolumn{2}{c|}{Clinc150}  & \multicolumn{2}{c|}{Mtop} & \multicolumn{2}{c|}{Massive} & \multicolumn{2}{c}{Average}\\
    & \textbf{NMI} & \textbf{Acc} & \textbf{NMI} & \textbf{Acc} & \textbf{NMI} & \textbf{Acc} & \textbf{NMI} & \textbf{Acc}  & \textbf{NMI} & \textbf{Acc}\\ \hline

  \rowcolor{gray!20} \textbf{Backbone embedder: MiniLM-L6-v2} & 79.08 & 61.27 & 89.38 & 73.59 & 67.29 & 31.19 & 70.24 & 52.80 & 76.50 & 54.71 \\

    GPT-4 \hspace{0.25cm} + IntentGPT  & \underline{81.42} & \underline{64.22} &  \underline{94.35} &  \underline{83.20} & - & -& - &  -& - & - \\
    GPT3.5 ~~ + IntentGPT & 76.58 & 54.82 &  90.04 &  73.68 & - & -& - &  -& - & -  \\
    \hline
    LLama ~~~ + SPILL  & 78.95 & 63.56 & 88.71 & 75.67 & 67.73 & 35.88 & 69.93 & 54.84 & 76.33 & 57.49 \\

    \hspace{1.05cm} + IDAS  &81.64 & 63.69 & 91.81 & 78.81 & 72.12 & 40.74 & 73.13 & 55.58 & 79.67 & 59.70 \\
    
    \hspace{1.05cm} + IDAS$\dagger$  & 80.82 & 62.48 & 91.64 & 81.72 & 71.52 & 39.74 & 74.06 & 59.77 & 79.51 & 60.93 \\

    \hspace{1.05cm} + LUMI (ours)  & \underline{82.78}$^*$ & \underline{67.33}$^*$ & \underline{93.21}$^*$ & \underline{83.13}$^*$ & \underline{73.86}$^*$ & \underline{46.49}$^*$ & \underline{74.94}$^*$ & \underline{61.77}$^*$ & \underline{81.20} & \underline{64.68} \\ \hline
    Gemma \hspace{0.1cm} + SPILL  & 81.92 & 66.22 & 90.78 & 77.76 & 70.48 & 37.95 & 72.59 & 58.05 & 78.94 & 60.00 \\

    \hspace{1.05cm} + IDAS  & 83.01 & 65.96 & 92.82 & 80.05 & 70.46 & 35.55 & 75.67 & 59.52 & 80.49 & 60.27 \\

    \hspace{1.05cm} + IDAS$\dagger$  & \textbf{\underline{84.88}} & \textbf{\underline{72.77}} & 93.41 & 85.52 & 70.88 & 36.46 & 75.93 & 57.75 & 81.27 & 63.13 \\
    
    \hspace{1.05cm} + LUMI (ours)   & 82.59 & 64.54 & \textbf{\underline{95.05}$^*$} & \textbf{\underline{86.93}$^*$} & \textbf{\underline{76.88}$^*$} & \textbf{\underline{49.10}$^*$} & \textbf{\underline{77.66}$^*$} & \textbf{\underline{64.18}$^*$} & \textbf{\underline{83.04}} & \textbf{\underline{66.19}} \\ \hline

    \rowcolor{gray!20} \textbf{Backbone embedder: Instructor-large} & 82.42 & 65.49 & 93.27 & 81.68 & 71.76 & 33.45 & 74.61 & 55.86 & 80.51 & 59.12 \\

   GPT3.5 ~~ + KeyphraseClust. & 82.4   &65.3  &  92.6  &79.4  &-&- &-&-& - & -\\

    \hspace{1.05cm} + ClusterLLM$\dagger$  &  \textbf{\underline{85.15}}  &\underline{71.20}   &  94.00  &83.80  & 73.83  &35.04   &  \underline{77.64}  & 60.69 & 82.66 & 62.68\\
   
    \hspace{1.05cm} + LLMEdgeRefine$\dagger$ & - & -  & \underline{94.86}  & \underline{86.77}  & \underline{72.92}  & \underline{46.00}  & 76.66  & \underline{63.42}  & - & -\\

    \hline

    LLama ~~~ + SPILL & 83.59 & \underline{69.93} & 92.90 & 84.59 & 71.11 & 35.26 & 75.56 & 60.06 & 80.79 & 62.46 \\

     \hspace{1.05cm} + IDAS  & 79.94 & 60.49 & 93.06 & 82.38 & 73.68 & 39.74 & 75.76 & 59.34 & 80.61 & 60.49 \\
    
     \hspace{1.05cm} + IDAS$\dagger$  & 81.79 & 67.45 & 92.12 & 81.71 & 71.69 & 37.25 & 76.88 & \underline{64.38} & 80.62 & 62.70 \\
   
     \hspace{1.05cm} + LUMI (ours)  & \underline{83.72}$^*$ & 67.12 & \underline{94.56}$^*$ & \underline{85.86}$^*$ & \underline{75.30}$^*$ & \underline{46.05}$^*$ & \underline{77.33}$^*$ & 63.91 & \underline{82.73} & \underline{65.73} \\

    \hline

    Gemma \hspace{0.1cm} +  SPILL  & \underline{85.01}  & 71.05  & 93.77   & 85.14   & 72.65   & 37.11  & 77.62  & 62.42  & 82.26 & 63.93\\

     \hspace{1.05cm} + IDAS  & 83.69 & 66.18 & 93.74 & 81.04 & 73.26 & 37.18 & 77.39 & 61.69 & 82.02 & 61.52 \\

     \hspace{1.05cm} + IDAS$\dagger$  & 84.45 & \textbf{\underline{71.56}} & 95.00 & \textbf{\underline{88.82}} & 72.24 & 38.28 & 78.04 & 61.89 & 82.43 & 65.14 \\

    \hspace{1.05cm} + LUMI (ours)   & 83.54 & 65.72 & \textbf{\underline{95.72}$^*$} & 87.31 & \textbf{\underline{79.11}$^*$} & \textbf{\underline{53.95}$^*$} & \textbf{\underline{78.88}$^*$} & \textbf{\underline{65.32}$^*$} & \textbf{\underline{84.31}} & \textbf{\underline{68.08}} \\

    \hline
  \end{tabular}

\end{table*}

\subsection{Analysis} \subsubsection{Ablation study} 
\begin{figure}[t]
    \centering

        \centering
        \includegraphics[width=0.95\columnwidth]{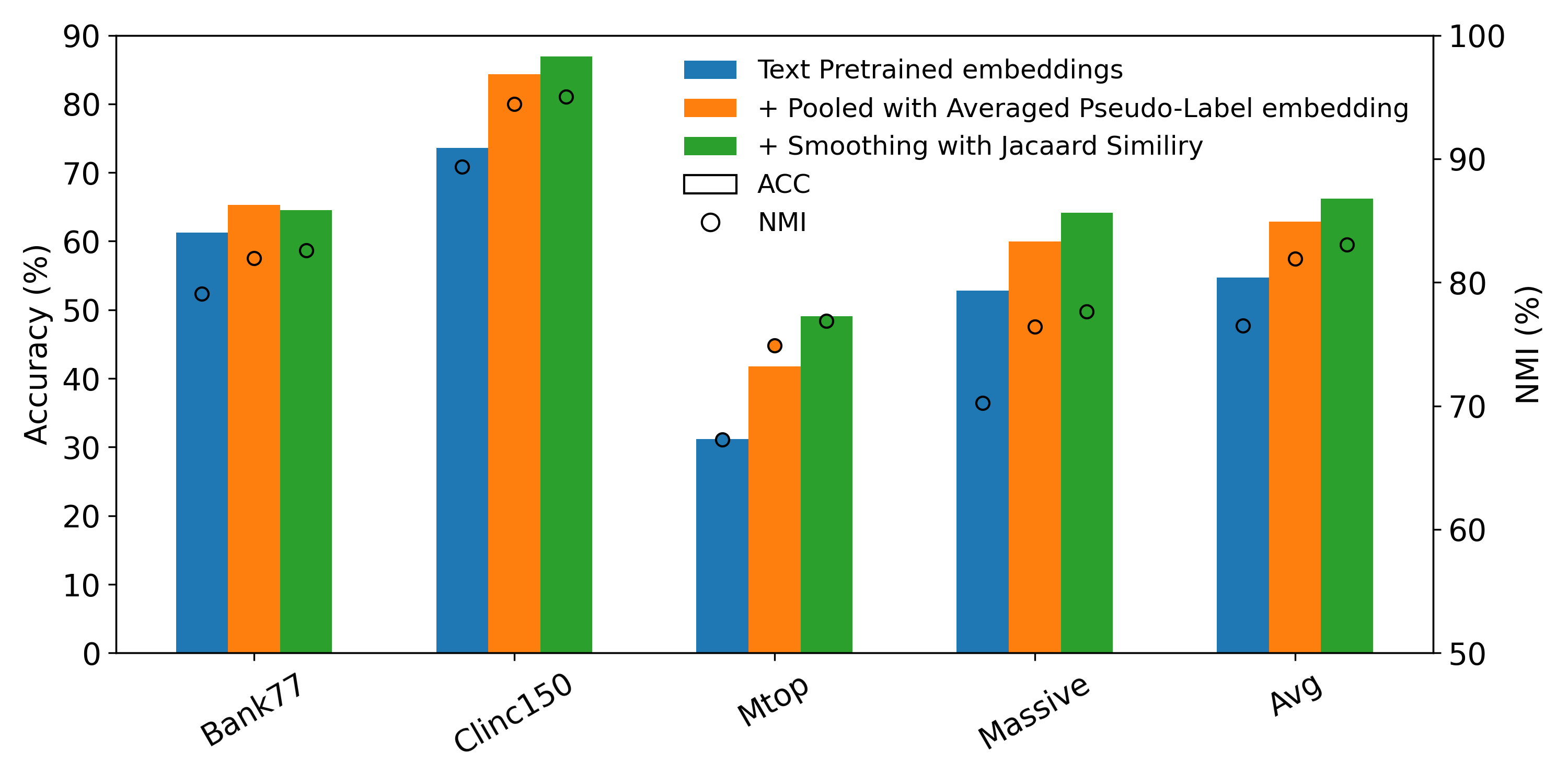}
        \caption{ 
    Cumulative contributions of method components: label-level and text-level pooling. 
    Embedder: MiniLM. LLM: Gemma. Averages over 5 runs. }
        \label{fig:componenet}

    \label{fig:combined}
\end{figure}

Figure \ref{fig:componenet} shows the cumulative effect of adding each method component. Overall, pooling text embeddings with averaged pseudo-label embedding leads to higher ACC and NMI compared to pretrained embeddings. Applying text-level pooling with Jaccard similarity weights further improves performance on three out of the four datasets.

\subsubsection{Variation of Candidate Count $b$} 
We set the top-10 budget ($b = 10$) for all experiments. To analyze the effect of candidate size, we vary $b \in {6, 8, 10, 12, 14}$ using MiniLM-L6-v2 and Gemma. All obtained NMI scores are between 82.4 and 83.7 and all Acc scores between 65.3 and 67.2, indicating that LUMI is relatively robust to variations in the candidate count.




\subsubsection{Error Analysis} \label{banking_error}
We hypothesize that Bank77's lower performance is due to pseudo-labels providing minimal denoising. To verify this, we analyze the results in detail (with MiniLM as embedder and Gemma as LLM). We find that, on average, each pseudo-label in Bank77 includes 1.70 ground truth clusters, while in CLINC150 it includes only 1.19, indicating that the pseudo-labels in Bank77 are noisier. Figure \ref{fig:componenet} also shows that the increase margin from averaged pseudo-label embeddings is the lowest compared to the others datasets. We manually checked the pseudo-labels and real labels. We found that the Banking intent is very fine-grained, e.g. ``I topped up my card but the money disappeared'' and ``My card was topped this morning but I can't see the funds. Why didn't it complete?'' have two different intents: `automatic top up' and `pending top up'. LUMI's pseudo-label `card top up issue' covers both. This suggests that few-shot examples may be needed for finer-grained intents. 

\section{Conclusion} 

We propose a novel method, LUMI, to improve text representations for short-text clustering using multiple pseudo-labels as guidance. LUMI does not rely on the impractical assumption of knowing the number of clusters $K$ and requires no labeled data to refine embeddings. We provide a theoretical analysis to explain why 
using multiple pseudo-labels improves performance. Our results show that LUMI overall outperform than SOTA methods across different LLMs and embedders. This indicates that our method is an effective training-free approach for unsupervised intent clustering. In the future, we plan to explore how our approach can be further improved on finer-grained datasets such as Bank77.

\bibliographystyle{ACM-Reference-Format}
\clearpage

\bibliography{references}

@inproceedings{hong2016accurate,
  title={Accurate and efficient query clustering via top ranked search results},
  author={Hong, Yuan and Vaidya, Jaideep and Lu, Haibing and Liu, Wen Ming},
  booktitle={Web Intelligence},
  volume={14},
  number={2},
  pages={119--138},
  year={2016},
  organization={SAGE Publications Sage UK: London, England}
}

@article{koskela2018proactive,
  title={Proactive information retrieval by capturing search intent from primary task context},
  author={Koskela, Markus and Luukkonen, Petri and Ruotsalo, Tuukka and Sj{\"o}berg, Mats and Flor{\'e}en, Patrik},
  journal={ACM Transactions on Interactive Intelligent Systems (TiiS)},
  volume={8},
  number={3},
  pages={1--25},
  year={2018},
  publisher={ACM New York, NY, USA}
}

@inproceedings{reimers-2019-sentence-bert,
  title = "Sentence-BERT: Sentence Embeddings using Siamese BERT-Networks",
  author = "Reimers, Nils and Gurevych, Iryna",
  booktitle = "Proceedings of the 2019 Conference on Empirical Methods in Natural Language Processing",
  month = "11",
  year = "2019",
  publisher = "Association for Computational Linguistics",
  url = "https://arxiv.org/abs/1908.10084",
}

@inproceedings{qu2019user,
  title={User intent prediction in information-seeking conversations},
  author={Qu, Chen and Yang, Liu and Croft, W Bruce and Zhang, Yongfeng and Trippas, Johanne R and Qiu, Minghui},
  booktitle={Proceedings of the 2019 Conference on Human Information Interaction and Retrieval},
  pages={25--33},
  year={2019}
}

@inproceedings{alexander2025few,
  title={In a Few Words: Comparing Weak Supervision and LLMs for Short Query Intent Classification},
  author={Alexander, Daria and de Vries, Arjen P},
  booktitle={Proceedings of the 48th International ACM SIGIR Conference on Research and Development in Information Retrieval},
  pages={2977--2981},
  year={2025}
}

@inproceedings{borvcin2024optimizing,
  title={Optimizing BERTopic: Analysis and reproducibility study of parameter influences on topic modeling},
  author={Bor{\v{c}}in, Martin and Jose, Joemon M},
  booktitle={European Conference on Information Retrieval},
  pages={147--160},
  year={2024},
  organization={Springer}
}

@inproceedings{liang2024survey,
  title={A Survey of Ontology Expansion for Conversational Understanding},
  author={Liang, Jinggui and Wu, Yuxia and Fang, Yuan and Fei, Hao and Liao, Lizi},
  booktitle={Proceedings of the 2024 Conference on Empirical Methods in Natural Language Processing},
  pages={18111--18127},
  year={2024}
}

@inproceedings{liang2024actively,
  title={Actively learn from llms with uncertainty propagation for generalized category discovery},
  author={Liang, Jinggui and Liao, Lizi and Fei, Hao and Li, Bobo and Jiang, Jing},
  booktitle={Proceedings of the 2024 Conference of the North American Chapter of the Association for Computational Linguistics: Human Language Technologies (Volume 1: Long Papers)},
  pages={7838--7851},
  year={2024}
}

@article{zou2025glean,
  title={GLEAN: Generalized Category Discovery with Diverse and Quality-Enhanced LLM Feedback},
  author={Zou, Henry Peng and Singh, Siffi and Nian, Yi and He, Jianfeng and Cai, Jason and Mansour, Saab and Su, Hang},
  journal={CoRR},
  year={2025}
}

@inproceedings{zhang2023clusterllm,
  title={ClusterLLM: Large Language Models as a Guide for Text Clustering},
  author={Zhang, Yuwei and Wang, Zihan and Shang, Jingbo},
  booktitle={Proceedings of the 2023 Conference on Empirical Methods in Natural Language Processing},
  pages={13903--13920},
  year={2023}
}

@inproceedings{rodriguez2024intentgpt,
  title={IntentGPT: Few-shot Intent Discovery with Large Language Models},
  author={Rodriguez, Juan A and Botzer, Nicholas and Vazquez, David and Pal, Christopher and Pedersoli, Marco and Laradji, Issam H},
  booktitle={ICLR 2024 Workshop on Large Language Model (LLM) Agents},
  year={2024}
}

@article{viswanathan-etal-2024-large,
    title = "Large Language Models Enable Few-Shot Clustering",
    author = "Viswanathan, Vijay  and
      Gashteovski, Kiril  and
      Gashteovski, Kiril  and
      Lawrence, Carolin  and
      Wu, Tongshuang  and
      Neubig, Graham",
    journal = "Transactions of the Association for Computational Linguistics",
    volume = "12",
    year = "2024",
    address = "Cambridge, MA",
    publisher = "MIT Press",
    url = "https://aclanthology.org/2024.tacl-1.18",
    doi = "10.1162/tacl_a_00648",
    pages = "321--333",
}

@inproceedings{lin2025spill,
    title = "{SPILL}: Domain-Adaptive Intent Clustering based on Selection and Pooling with Large Language Models",
    author = "Lin, I-Fan  and
      Hasibi, Faegheh  and
      Verberne, Suzan",
    editor = "Che, Wanxiang  and
      Nabende, Joyce  and
      Shutova, Ekaterina  and
      Pilehvar, Mohammad Taher",
    booktitle = "Findings of the Association for Computational Linguistics: ACL 2025",
    month = jul,
    year = "2025",
    address = "Vienna, Austria",
    publisher = "Association for Computational Linguistics",
    url = "https://aclanthology.org/2025.findings-acl.812/",
    doi = "10.18653/v1/2025.findings-acl.812",
    pages = "15723--15737",
    ISBN = "979-8-89176-256-5",
}

@inproceedings{de2023idas,
  title={IDAS: Intent Discovery with Abstractive Summarization},
  author={De Raedt, Maarten and Godin, Fr{\'e}deric and Demeester, Thomas and Develder, Chris},
  booktitle={Proceedings of the 5th Workshop on NLP for Conversational AI (NLP4ConvAI 2023)},
  pages={71--88},
  year={2023}
}

@article{viswanathan2024large,
  title={Large Language Models Enable Few-Shot Clustering},
  author={Viswanathan, Vijay and Gashteovski, Kiril and Lawrence, Carolin and Wu, Tongshuang and Neubig, Graham},
  journal={Transactions of the Association for Computational Linguistics},
  volume={11},
  pages={321--333},
  year={2024}
}

@inproceedings{feng2024llmedgerefine,
  title={LLMEdgeRefine: Enhancing text clustering with LLM-based boundary point refinement},
  author={Feng, Zijin and Lin, Luyang and Wang, Lingzhi and Cheng, Hong and Wong, Kam-Fai},
  booktitle={Proceedings of the 2024 Conference on Empirical Methods in Natural Language Processing},
  pages={18455--18462},
  year={2024}
}

@inproceedings{larson2019evaluation,
  title={An Evaluation Dataset for Intent Classification and Out-of-Scope Prediction},
  author={Larson, Stefan and Mahendran, Anish and Peper, Joseph J and Clarke, Christopher and Lee, Andrew and Hill, Parker and Kummerfeld, Jonathan K and Leach, Kevin and Laurenzano, Michael A and Tang, Lingjia and others},
  booktitle={Proceedings of the 2019 Conference on Empirical Methods in Natural Language Processing and the 9th International Joint Conference on Natural Language Processing (EMNLP-IJCNLP)},
  pages={1311--1316},
  year={2019}
}

@inproceedings{Casanueva2020,
    author      = {I{\~{n}}igo Casanueva and Tadas Temcinas and Daniela Gerz and Matthew Henderson and Ivan Vulic},
    title       = {Efficient Intent Detection with Dual Sentence Encoders},
    year        = {2020},
    month       = {mar},
    note        = {Data available at https://github.com/PolyAI-LDN/task-specific-datasets},
    url         = {https://arxiv.org/abs/2003.04807},
    booktitle   = {Proceedings of the 2nd Workshop on NLP for ConvAI - ACL 2020}
}

@inproceedings{fitzgerald2023massive,
  title={MASSIVE: A 1M-Example Multilingual Natural Language Understanding Dataset with 51 Typologically-Diverse Languages},
  author={Fitzgerald, Jack and Hench, Christopher and Peris, Charith and Mackie, Scott and Rottmann, Kay and Sanchez, Ana and Nash, Aaron and Urbach, Liam and Kakarala, Vishesh and Singh, Richa and others},
  booktitle={Proceedings of the 61st Annual Meeting of the Association for Computational Linguistics (Volume 1: Long Papers)},
  pages={4277--4302},
  year={2023}
}

@article{li2020mtop,
  title={MTOP: A comprehensive multilingual task-oriented semantic parsing benchmark},
  author={Li, Haoran and Arora, Abhinav and Chen, Shuohui and Gupta, Anchit and Gupta, Sonal and Mehdad, Yashar},
  journal={arXiv preprint arXiv:2008.09335},
  year={2020}
}

@article{riviere2024gemma,
  title={Gemma 2: Improving Open Language Models at a Practical Size},
  author={Rivi{\`e}re, Morgane and Pathak, Shreya and Sessa, Pier Giuseppe and Hardin, Cassidy and Bhupatiraju, Surya and Hussenot, L{\'e}onard and Mesnard, Thomas and Shahriari, Bobak and Ram{\'e}, Alexandre and Ferret, Johan and others},
  journal={CoRR},
  year={2024}
}

@article{llama3modelcard,
  title={Llama 3 Model Card},
  author={AI@Meta},
  year={2024},
  url = {https://github.com/meta-llama/llama3/blob/main/MODEL_CARD.md}
}

@inproceedings{su2023one,
  title={One Embedder, Any Task: Instruction-Finetuned Text Embeddings},
  author={Su, Hongjin and Shi, Weijia and Kasai, Jungo and Wang, Yizhong and Hu, Yushi and Ostendorf, Mari and Yih, Wen-tau and Smith, Noah A and Zettlemoyer, Luke and Yu, Tao},
  booktitle={Findings of the Association for Computational Linguistics: ACL 2023},
  pages={1102--1121},
  year={2023}
}

@article{meilua2007comparing,
  title={Comparing clusterings—an information based distance},
  author={Meil{\u{a}}, Marina},
  journal={Journal of multivariate analysis},
  volume={98},
  number={5},
  pages={873--895},
  year={2007},
  publisher={Elsevier}
}

@inproceedings{huang2014deep,
  title={Deep embedding network for clustering},
  author={Huang, Peihao and Huang, Yan and Wang, Wei and Wang, Liang},
  booktitle={2014 22nd International conference on pattern recognition},
  pages={1532--1537},
  year={2014},
  organization={IEEE}
}

@inproceedings{gung2023intent,
  title={Intent Induction from Conversations for Task-Oriented Dialogue Track at DSTC 11},
  author={Gung, James and Shu, Raphael and Moeng, Emily and Rose, Wesley and Romeo, Salvatore and Gupta, Arshit and Benajiba, Yassine and Mansour, Saab and Zhang, Yi},
  booktitle={Proceedings of The Eleventh Dialog System Technology Challenge},
  pages={242--259},
  year={2023}
}

@article{rand1971objective,
  title={Objective criteria for the evaluation of clustering methods},
  author={Rand, William M},
  journal={Journal of the American Statistical association},
  volume={66},
  number={336},
  pages={846--850},
  year={1971},
  publisher={Taylor \& Francis}
}

\end{document}